\pdfoutput=1

\documentclass[11pt]{article}
\usepackage[]{eacl2023}

\usepackage{times}
\usepackage{latexsym}

\usepackage[T1]{fontenc}
\usepackage[utf8]{inputenc}
\usepackage{graphicx}
\usepackage{multirow}
\usepackage{float}
\usepackage{cleveref}
\usepackage{microtype}
\everypar{looseness=-1}

\usepackage[suppress]{color-edits}
\addauthor{ZT}{olive} 
\addauthor{JG}{red} 
\addauthor{DG}{cyan} 
\addauthor{JM}{orange} 

\usepackage{color, colortbl}
\definecolor{LightCyan}{rgb}{0.88,1,1}
\definecolor{Gray}{gray}{0.9}

\title{A Federated Approach for Hate Speech Detection}

\author{Jay Gala* \\
  AI4Bharat\\
  \texttt{jaygala24@gmail.com} \\\And
  Deep Gandhi*\\
  University of Alberta \\
  \texttt{drgandhi@ualberta.ca} \\\AND
  Jash Mehta*\\
  Georgia Institute of Technology\\
  \texttt{jmehta73@gatech.edu} \\\And
  Zeerak Talat\\
  MBZUAI\\
  }

\begin{document}
\maketitle
\def\thefootnote{*}\footnotetext{Equal contribution.}\def\thefootnote{\arabic{footnote}}
\begin{abstract}
Hate speech detection has been the subject of high research attention, due to the scale of content created on social media.
In spite of the attention and the sensitive nature of the task, privacy preservation in hate speech detection has remained under-studied.
The majority of research has focused on centralised machine learning infrastructures which risk leaking data.
In this paper, we show that using federated machine learning can help address privacy the concerns that are inherent to hate speech detection while obtaining up to $6.81\%$ improvement in terms of F1-score.
\end{abstract}

\section{Introduction}
Content moderation is a topic that intersects across multiple fundamental rights, e.g., freedom of expression and the right to privacy; and interest groups, e.g. scholars, legislators, civil society, and commercial entities \cite{Kaye_2019}.
The availability of public datasets has been crucial to the development of computational methods for hate speech detection.
However, public data contains risks for those whose content is available.
On the other hand, privately held data, e.g., data held by corporate entities, holds risks for those who are reporting content.
Such risks may be actualised through information leaks in models \cite{hitaj2017deep} or the transmission of data \cite{shokri2015privacy}, and can impact people's safety and livelihood.

\begin{figure}[t]
\centering
\includegraphics[width=\columnwidth]{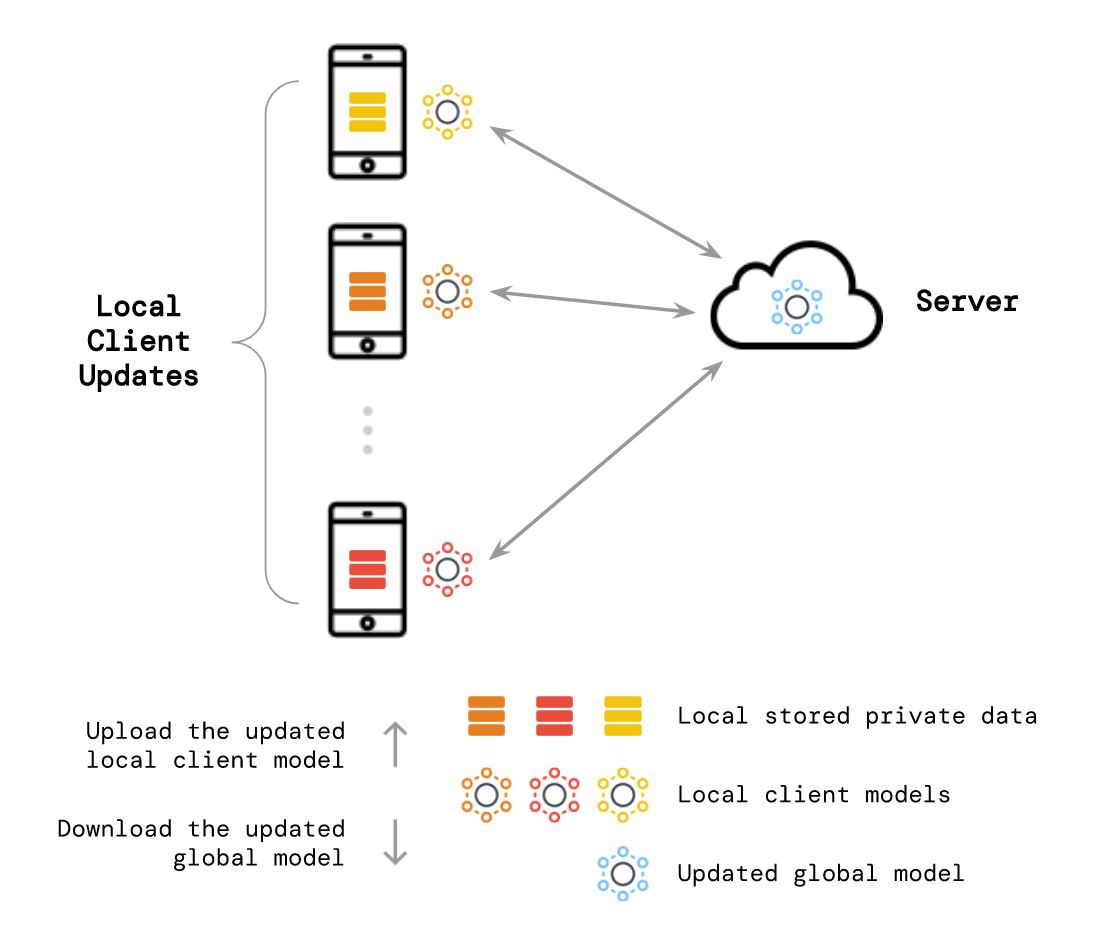}
\caption{Federated Learning: A centralised model is hosted on a server and is distributed to client devices, these compute weight updates, and transmit the updates for aggregation into the centralised model.
The centralised model is then redistributed to client devices.}
\label{fig:federated-learning}
\vspace{-0.45cm}
\end{figure}

In this work, we apply Federated Learning \cite[FL,][]{mcmahan2017communication} to address the lack of privacy in hate speech detection.
FL is a privacy-preserving training paradigm for machine learning that jointly optimises for user privacy and model performance.
We posit that privacy is necessary for users whose content is flagged and users who are flagging content alike.
We thus operationalise privacy, in the context of hate speech detection and federated learning, to mean privacy in terms of the content of reported content, and the report itself.
FL is an apt training paradigm for \ZTdelete{hate speech detection and other} tasks in which training data is highly sensitive, as FL is designed to mitigate risks of information leaks while also dealing with a high number of end-users, information loss, and label imbalances \cite{lin-etal-2022-fednlp,priyanshu2021fedpandemic,gandhi-etal-2022-federated}.
We apply the FL algorithms FedProx \citep{li2018federated} and Adaptive Federated Optimization \cite[FedOpt,][]{reddi_2021_fedopt} to $5$ machine learning algorithms.
We evaluate our approach on $8$ previously published datasets for hate speech detection.
While using FL often implies a trade-off between privacy and performance, we obtain performance improvements of up to $6.81\%$ in F1-score.
\ZTdelete{that the use of FL can provide performance boosts of up to $6.81\%$ in F1-score.}
We find that that models trained using FL outperform centralised models across multiple tests (e.g., derogatory language, spelling variation, and pronoun reference) in \textsc{HateCheck} \cite{rottger-etal-2021-hatecheck}.\footnote{\DGedit{All code is made available on \href{https://github.com/jaygala24/fed-hate-speech}{Github}}.}
\ZTdelete{Analysing model performances on \textsc{HateCheck} \cite{rottger-etal-2021-hatecheck}, we find that models trained using FL outperform centralised models across multiple categories (e.g., derogatory language, spelling variation, and pronoun reference).}\ZTdelete{in terms of derogatory language, pronoun reference, phrasing, spelling variation, and threatening language.}
\ZTdelete{In an analysis using \textsc{HateCheck \cite{rottger-etal-2021-hatecheck}, we find that the} FL models outperform the centralised models in terms of derogatory language, pronoun reference, phrasing, spelling variations, and threatening language.}

\section{Prior Work}
Although the areas of hate speech detection and FL have each been subject to extensive research, the study of their intersection remains in its infancy.

\paragraph{Federated Learning}
Federated Learning is a privacy-preserving machine learning paradigm that aims to reduce privacy risks by decentralising data processing onto client devices (i.e., personal devices), thereby foregoing the need for transmitting ``raw'' user data, and thus minimising risks of personal data leaks caused by transmission of data.\footnote{See \newcite{gitelman_raw1_2013} for a discussion on `raw' data.}
In FL, the machine learning model is located in two places: On a centralised server, and on client devices, which hold instances of the model distributed from the centralised model.
Client devices use the model to compute model updates.
The model updates are then transmitted to the server and aggregated by the centralised model, which is redistributed to the client devices.
However, not all transmitted weight updates are aggregated into the model.
FL operates with a notion of data loss in its design, which is emulated by selecting a fraction of clients whose updates are aggregated.
Thus, FL paradigm uses less data to train a models.

In our experiments, we apply two FL algorithms: FedProx and FedOpt \cite{reddi_2021_fedopt}.
FedProx introduces a proximal term to  the Federated Averaging algorithm \cite[FedAvg,][]{mcmahan2017communication}.
FedAvg averages the weights computed on participating client devices in a round.
FedProx introduces a proximal term that functions as a regulariser to the weight updates transmitted by participating clients, which penalises local weight updates that diverge from the global model.
The FedAvg algorithm can thus be understood as a special case of FedProx with the proximal term set to $0.0$.

\JGedit{FedOpt \citep{reddi_2021_fedopt} extends the adaptive optimisation strategies from centralised optimisation (e.g., Adam \cite{adam_2014} and Adagrad \cite{adagrad_2010}) to explicitly account for client and server optimisation. 
\ZTedit{FedOpt handles server optimisation distinctly from client optimisation, by introducing a state to the server-side optimisation routine.}
\ZTdelete{In this algorithm, the client optimization is treated fundamentally different from the server optimization, where the server maintains a state throughout the federated optimization.} 
\DGdelete{The server-side adaptivity enables more accurate and heterogeneity-aware FL models, thereby offering favorable convergence \JGcomment{for non-convex settings}}.
\DGedit{This distinct handling of server-side optimization enables more accurate and heterogeneity-aware FL models, which can speed up convergence.}
}

FL has been applied to a number of tasks, including emoji prediction \cite{ramaswamy2019federated,gandhi-etal-2022-federated}, next-word prediction for mobile keyboards \cite{yang2018applied}, pre-training and fine-tuning large language models \cite{liu2020federated}, medical named entity recognition \cite{ge2020fedner}, and text classification \cite{lin-etal-2022-fednlp}.
For instance, \newcite{lin-etal-2022-fednlp} used FL to fine-tune a DistilBERT model to perform classification on the 20NewsGroup dataset \cite{Lang95} using three different FL algorithms: FedAvg~\citep{mcmahan2017communication}, FedProx~\citep{li2018federated}, FedOpt~\citep{reddi_2021_fedopt}) under non-IID partitioning.

In a closely related study, \newcite{Basu2021BenchmarkingDP} apply FL, using the FedAvg algorithm to fine-tune large language models to detect depression and sexual harassment from small Twitter data samples.
They find that using large language models such as BERT and RoBERTa outperform distilled language models such as DistilBERT.
Our work extends on \newcite{Basu2021BenchmarkingDP} by introducing additional FL algorithms and extending to a multi-class setting for hate speech detection.
\ZTdelete{experimenting with a multi-class setting for hate speech detection and introducing more FL algorithms.\ZTdelete{to hate speech detection.the FedProx and FedOpt algorithms to hate speech detection.}}

Thus, our work extends on prior work by i) applying FL to the task of multi-class hate speech detection, a task which has proven difficult in part due to the complex nature of pragmatics \cite{rottger-etal-2021-hatecheck} and hate mongers seeking to evade content moderation infrastructures \cite{crawford_what_2016}; ii) using the FedProx and FedOpt algorithms rather than the FedAvg algorithm, thereby reducing model vulnerability to divergent weight updates; and iii) providing an in-depth analysis of federated model performances.

\paragraph{Hate Speech Detection}

Prior work on hate speech detection has primarily focused on privacy-agnostic machine learning paradigms, using centralised models for classification.
Such work has investigated a number of machine learning models (e.g. SVMs \cite{karan_cross-domain_2018}, CNNs \cite{park-fung-2017-one}, and fine-tuned language models \citep{swamy_studying_2019}) and the development of resources \cite[e.g.][]{talat_hateful_2016}.
Recently, \newcite{fortuna2021well} proposed a standardisation of classes across $9$ publicly available datasets and studied the generalisation capabilities of BERT, fastText, and SVM models. \ZTdelete{experimented with BERT, fastText, and SVM models to study their ability to generalise.} 
In their work they found limited success in inter-dataset generalization.
Our work thus extends on the task of hate speech detection by introducing privacy-preserving methods to multi-class hate speech detection.
In doing so, the privacy of those who flag content and those whose content is flagged remain intact.

\begin{table}[!h]
\centering
\resizebox{\columnwidth}{!}{
\begin{tabular}{c|rrr}
{\textbf{Category}}  &
{\textbf{Merged count}} &
{\texttt{{\bf Comb}}} & \textbf{Change}\\\hline

aggression             & $6,950$   & $6,950$ & -  \\
aggressive hate speech & $1,561$   & $1,561$ & - \\
covert aggression      & $4,242$   & $4,242$ & - \\
\textit{hate speech}   & $13,222$  & $13,205$ & $-0.13\%$  \\
\textit{insult}        & $7,879$   & $7,779$ & $-1.27\%$    \\
misogyny sexism        & $5,000$   & $5,000$ & -\\
\textit{none}          & $189,869$ & $188,550$ & $-0.69\%$ \\
offensive              & $19,192$  & $19,192$ & - \\
overt aggression       & $2,710$   & $2,710$ & -     \\
racism                 & $1,978$   & $1,978$ & - \\
\textit{severely toxic}  & $1,597$   & $1,527$ & $-4.38\%$    \\
\textit{threat}        & $480$     & $470$ & $-2.08\%$   \\
\textit{toxicity}      & $40,316$  & $40,134$ & $-0.45\%$   \\
\end{tabular}
}
\caption{Label count of the raw datasets and \texttt{Comb}}
\label{category-table-combined}
\end{table}

\begin{table*}[!ht]

\resizebox{\textwidth}{!}{
\begin{tabular}{cc|ccc|ccc|ccc|ccc|ccc}\\
 &
   &
  \multicolumn{3}{c|}{\textbf{Logistic Regression}} &
  \multicolumn{3}{c|}{\textbf{Bi-LSTM}} &
  \multicolumn{3}{c|}{\textbf{FNet}} &
  \multicolumn{3}{c|}{\textbf{DistilBERT}} &
  \multicolumn{3}{c}{\textbf{RoBERTa}} \\
 &
   &
  Precision &
  Recall &
  F1 &
  Precision &
  Recall &
  F1 &
  Precision &
  Recall &
  F1 &
  Precision &
  Recall &
  F1  &
  Precision &
  Recall &
  F1\\\hline
\multirow{3}{*}{\textbf{c = $10\%$}} &
  \textbf{e = 1} &
  $70.22$ &
  $53.15$ &
  $58.47$ &
  $71.04$ &
  $58.19$ &
  $61.28$ &
  $72.61$ &
  $59.20$ &
  $62.20$ &
  $73.98$ &
  $60.75$ &
  $63.79$ &
  $74.76$ &
  $64.43$ &
  $66.16$ \\ 
 &
 
  \textbf{e = 5} &
  $70.83$ &
  $63.31$ &
  $66.35$ &
  $70.84$ &
  $66.51$ &
  $67.72$ &
  $73.52$ &
  $68.33$ &
  $\textbf{70.42}$ &
  $74.54$ &
  $69.46$ &
  $70.85$ &
  $74.59$ &
  $69.68$ &
  $71.48$ \\ 
 &
  \textbf{e = 20} &
  $70.18$ &
  $67.41$ &
  $68.67$ &
  $69.17$ &
  $69.25$ &
  $69.10$ &
  $73.10$ &
  $68.02$ &
  $69.73$ &
  $73.28$ &
  $71.06$ &
  $71.94$ &
  $73.11$ &
  $\textbf{71.48}$ &
  $\textbf{72.07}$ \\\hline
\multirow{3}{*}{\textbf{c = 30\%}} &

  \textbf{e = 1} &
  $\textbf{71.23}$ &
  $53.50$ &
  $58.89$ &
  $71.58$ &
  $58.82$ &
  $61.72$ &
  $73.62$ &
  $61.13$ &
  $63.97$ &
  $74.84$ &
  $64.03$ &
  $66.14$ &
  $75.02$ &
  $64.33$ &
  $66.41$ \\ 
 &
  \textbf{e = 5} &
  $70.82$ &
  $64.44$ &
  $67.01$ &
  $70.65$ &
  $65.90$ &
  $67.27$ &
  $73.35$ &
  $68.30$ &
  $70.36$ &
  $74.82$ &
  $69.44$ &
  $70.68$ &
  $74.41$ &
  $69.98$ &
  $71.81$ \\ 
 &
 
  \textbf{e = 20} &
  $70.30$ &
  $\textbf{68.13}$ &
  $\textbf{69.09}$ &
  $69.34$ &
  $\textbf{69.26}$ &
  $\textbf{69.15}$ &
  $72.35$ &
  $68.03$ &
  $69.74$ &
  $73.33$ &
  $71.39$ &
  $72.15$ &
  $73.65$ &
  $70.86$ &
  $71.96$ \\\hline
\multirow{3}{*}{\textbf{c = 50\%}} &
  \textbf{e = 1} &
  $71.11$ &
  $53.12$ &
  $58.58$ &
  $\textbf{71.59}$ &
  $58.71$ &
  $61.73$ &
  $\textbf{73.93}$ &
  $61.89$ &
  $64.51$ &
  $\textbf{74.88}$ &
  $63.58$ &
  $65.85$ &
  $74.42$ &
  $63.57$ &
  $65.87$ \\ 
 &
 
  \textbf{e = 5} &
  $70.89$ &
  $64.26$ &
  $66.80$ &
  $70.70$ &
  $66.16$ &
  $67.54$ &
  $72.90$ &
  $68.27$ &
  $70.18$ &
  $74.44$ &
  $69.68$ &
  $70.88$ &
  $\textbf{74.90}$ &
  $69.46$ &
  $70.86$ \\ 
 &
 \textbf{e = 20} &
  $70.28$ &
  $68.00$ &
  $69.01$ &
  $69.25$ &
  $68.84$ &
  $68.20$ &
  $72.90$ &
  $\textbf{68.42}$ &
  $70.16$ &
  $73.71$ &
  $\textbf{71.51}$ &
  $\textbf{72.34}$ &
  $73.53$ &
  $71.18$ &
  $72.01$
\end{tabular}
}
\caption{Results of FedProx experiments on \texttt{Comb}.}
\label{tab:fed-combined-results}

\end{table*}

\begin{table*}[!ht]

\resizebox{\textwidth}{!}{
\begin{tabular}{cc|ccc|ccc|ccc|ccc|ccc}
 &
   &
  \multicolumn{3}{c|}{\textbf{Logistic Regression}} &
  \multicolumn{3}{c|}{\textbf{Bi-LSTM}} &
  \multicolumn{3}{c|}{\textbf{FNet}} &
  \multicolumn{3}{c|}{\textbf{DistilBERT}} &
  \multicolumn{3}{c}{\textbf{RoBERTa}} \\
 &
   &
  Precision &
  Recall &
  F1 &
  Precision &
  Recall &
  F1 &
  Precision &
  Recall &
  F1 &
  Precision &
  Recall &
  F1  &
  Precision &
  Recall &
  F1\\\hline
\multirow{3}{*}{\textbf{c = 10\%}} &
  \textbf{e = 1} &
  $68.29$ &
  $52.48$ &
  $58.46$ &
  $\textbf{71.80}$ &
  $58.34$ &
  $61.70$ &
  $72.64$ &
  $59.78$ &
  $62.49$ &
  $74.51$ &
  $63.87$ &
  $65.03$ &
  $72.02$ &
  $64.21$ &
  $64.57$ \\ 
 &
 
  \textbf{e = 5} &
  $68.20$ &
  $59.38$ &
  $63.14$ &
  $70.63$ &
  $63.73$ &
  $66.10$ &
  $72.39$ &
  $69.11$ &
  $70.51$ &
  $74.33$ &
  $69.61$ &
  $70.48$ &
  $\textbf{75.55}$ &
  $69.44$ &
  $70.34$ \\ 
 &
  \textbf{e = 20} &
  $\textbf{68.30}$ &
  $59.56$ &
  $63.23$ &
  $69.74$ &
  $65.32$ &
  $67.27$ &
  $71.87$ &
  $\textbf{70.69}$ &
  $\textbf{71.15}$ &
  $72.21$ &
  $\textbf{71.34}$ &
  $\textbf{71.66}$ &
  $73.17$ &
  $\textbf{72.20}$ &
  $\textbf{72.61}$ \\\hline
\multirow{3}{*}{\textbf{c = 30\%}} &

  \textbf{e = 1} &
  $67.68$ &
  $51.05$ &
  $57.19$ &
  $71.56$ &
  $58.97$ &
  $62.10$ &
  $72.24$ &
  $57.11$ &
  $61.21$ &
  $\textbf{74.90}$ &
  $64.16$ &
  $66.79$ &
  $73.88$ &
  $66.07$ &
  $65.79$ \\ 
 &
  \textbf{e = 5} &
  $66.65$ &
  $60.31$ &
  $63.10$ &
  $69.48$ &
  $62.63$ &
  $65.57$ &
  $72.01$ &
  $69.14$ &
  $70.30$ &
  $72.82$ &
  $69.59$ &
  $70.75$ &
  $74.38$ &
  $71.54$ &
  $71.69$ \\ 
 &
 
  \textbf{e = 20} &
  $67.18$ &
  $62.50$ &
  $\textbf{64.60}$ &
  $69.74$ &
  $65.69$ &
  $67.49$ &
  $71.91$ &
  $70.02$ &
  $70.79$ &
  $71.55$ &
  $70.33$ &
  $70.86$ &
  $72.97$ &
  $72.10$ &
  $72.05$ \\\hline
\multirow{3}{*}{\textbf{c = 50\%}} &
  \textbf{e = 1} &
  $67.25$ &
  $54.85$ &
  $59.82$ &
  $71.35$ &
  $59.63$ &
  $62.59$ &
  $\textbf{73.03}$ &
  $62.28$ &
  $64.64$ &
  $73.31$ &
  $63.98$ &
  $65.64$ &
  $74.85$ &
  $66.80$ &
  $67.75$ \\ 
 &
  \textbf{e = 5} &
  $66.63$ &
  $60.21$ &
  $63.04$ &
  $69.56$ &
  $63.02$ &
  $65.58$ &
  $70.63$ &
  $68.06$ &
  $69.21$ &
  $72.53$ &
  $69.66$ &
  $70.80$ &
  $73.78$ &
  $71.27$ &
  $71.18$ \\ 
 &
 \textbf{e = 20} &
  $66.70$ &
  $\textbf{62.51}$ &
  $64.41$ &
  $69.16$ &
  $\textbf{66.16}$ &
  $\textbf{67.54}$ &
  $70.98$ &
  $69.74$ &
  $70.21$ &
  $70.65$ &
  $68.99$ &
  $69.69$ &
  $72.67$ &
  $70.51$ &
  $71.51$ \\
\end{tabular}
}
\caption{Results of FedOpt experiments on \texttt{Comb}.}
\label{tab:fedopt-combined-results}

\end{table*}

\begin{table}[!ht]

\resizebox{\columnwidth}{!}{
\centering
\begin{tabular}{lccc|c}

& \multicolumn{3}{c|} {\textbf{Centralised}}
& {\textbf{Federated}}\\

 & Precision & Recall & F1 & F1 
\\\hline

LogReg
& $69.11$ & $57.45$ & $62.20$ & $69.09$

\\

Bi-LSTM 
& $71.43$ & $66.64$ & $67.90$ & $69.15$

\\

FNet 
& $71.35$ & $64.73$ & $66.58$ & $71.15$

\\

DistilBERT 
& $73.99$ & $69.01$ & $69.39$ & $72.34$ 

\\

RoBERTa 
& $\mathbf{75.45}$ & $\mathbf{70.58}$ & $\mathbf{71.03}$ & $\mathbf{72.61}$\\

\end{tabular}
}

\caption{Results for the centralised and best performing FL models. \DGedit{The FL models have been chosen across FedProx and FedOpt based on F1 scores.}}
\label{tab:server-results}
\vspace{-0.3cm}
\end{table}

\section{Data}
\label{sec:dataset}
We combine our dataset using the standardisation schema proposed by \newcite{fortuna2021well}.

\paragraph{\texttt{Comb}}
\label{combined-dataset}
We reuse $8$ of the $9$ datasets used by \newcite{fortuna2021well} to form \texttt{Comb}.\footnote{The dataset proposed by \cite{founta2018large} is not included as it was not available to us.}
\texttt{Comb} then consists of the datasets proposed by \newcite{talat_hateful_2016,davidson_automated_2017,fersini2018overview,de-gibert-etal-2018-hate,swamy_studying_2019,basile-etal-2019-semeval,zampieri-etal-2019-predicting} and the Kaggle toxic comment challenge.\footnote{\label{kaggle-data}https://www.kaggle.com/c/jigsaw-toxic-comment-classification-challenge}
We perform a stratified split of all training data into training (70\%), validation (10\%), and test (20\%) sets.\footnote{We do not use the test data provided with some datasets to ensure uniformity, as test sets are not provided with all datasets.}

\paragraph{Data Cleaning} 
We address issues of extreme class imbalance in \texttt{Comb} by removing the ``abusive'' category as it only contains $2$ documents.
Following an in-depth analysis of the \texttt{Kaggle} dataset we find that the maximum length of tokens in the dataset is $4950$ while the median length of tokens in \texttt{Comb} is $26$.
Moreover, we find that the longest $1\%$ of documents in the \texttt{Kaggle} dataset do not contain unique tokens.
Removing the longest $1\%$ of comments reduces the maximal document length to $727$ tokens (see \Cref{sec:appendix-token-dist} for further detail).
Following our data cleaning processes, \texttt{Comb} comes to consist of $293,300$ documents (see \Cref{category-table-combined} for an overview of changes).

\section{Experiments}
\label{sec:experiments}
We experiment with $5$ machine learning models in their centralised and federated settings: Logistic Regression Bi-LSTMs \citep{Hochreiter_Schmidhuber_1997}, FNet \citep{lee-thorp-etal-2022-fnet}, DistilBERT \citep{sanh2019distilbert} and RoBERTa \citep{liu2019roberta}. 
We measure their performance using weighted F1 scores.
The centralised models form our baselines, while the federated models form our experimental models.
For the Logistic Regression and Bi-LSTMs, we perform word-level tokenisation using SpaCy \citep{honnibal_spacy1_2017}.
For the FNet, DistilBERT, and RoBERTa, we use the tokenisers provided with each model.\footnote{Please refer to \Cref{sec:appendix} for further experiments and analyses on the \newcite{vidgen-etal-2021-learning} dataset.}

\subsection{Federated Training}
FL is a machine learning training paradigm that distributes training onto client devices.
All client devices are split into overlapping subsets and the training data is partitioned and uniformly distributed to client devices.
A random client subset is selected for training in each round, and their locally computed weights are aggregated on the server.
We train our models for $300$ rounds for $1$, $5$, or $20$ epochs per round, and set the client fraction to $10\%$, $30\%$, or $50\%$ which are randomly sampled from $100$ client devices.
We perform hyper-parameter tuning for the client learning rate, server-side learning rate, and proximal term (see \cref{app:hyper-params}).

In our work, we conceptualise client devices as users who witness and report hate speech.
We simulate the client devices and ensure that data is independently and identically distributed (I.I.D.) on client devices.\footnote{We use an I.I.D. setting for data as $40\%$ of all social media users and $64\%$ of those under 30 in the USA have experienced online harassment \cite{pew_research_center_state_2021}.
I.e. while hate speech is not frequent, it is often experienced by users.}
We use the FedProx and FedOpt algorithms to aggregate client updates on the server.
FedProx introduces a regularisation constant to the server-side aggregation step, the proximal term to address issues of divergence in weights and statistical heterogeneity in FedAvg.
FedOpt seeks to create more robust models by introducing a separate optimiser for the server-side model to account for data heterogeneity.

\section{Analysis}
\label{sec:main-analysis}
Considering the baseline models in \Cref{tab:server-results}, we see that the Logistic Regression tends to under-perform, while the RoBERTa model posts the best performances.
Although FL-based models often outperform our baselines, we note that when FL models are trained with lower client fractions and epochs, they tend to be outperformed by the baselines.
Models trained using FedProx outperform the centralised baselines (see \cref{tab:fed-combined-results}).\footnote{For tables \ref{tab:fed-combined-results} and \ref{tab:fedopt-combined-results}, $c$ refers to the client fraction used and $e$ refers to the number of epochs on client devices.}
For instance, we see large improvements for FNet and Logistic Regression ($4.5$ and $6.8$ points in terms of F1- score, respectively).
Comparing the performances of models trained using FedOpt (\cref{tab:fedopt-combined-results}) with those trained using FedProx, we observe that the former (in particular FNet and RoBERTa) tend to outperform the latter for lower client fractions and epochs.
In general, we find that the best FL models outperform their centralised counter-parts (see Tables \ref{tab:fed-combined-results} and \ref{tab:fedopt-combined-results}). 
In fact, the best performing RoBERTa, DistilBERT, and FNet models trained using FL algorithms outperform their centralised baselines, with FNet obtaining a $3$-$4$ point improvement over centralised models in terms of F1 score.\footnote{See \Cref{app:hate-check} for an analysis using HateCheck \cite{rottger-etal-2021-hatecheck}.}

While FL often indicates a trade-off between privacy and performance, we find that the best FL models outperform the centralised baselines.
We believe that the improved performance stems from the dataset being split into smaller segments, in congruence with findings from prior work.
For instance, \newcite{nobata_2016_smaller_units} show that splitting data into smaller temporal segments helped improve classification performance overall.
We believe that a similar effect may be evident with FL models that, by design split data into small segments and disregard a fraction of the clients. 
Further, it may be the case that some data within hate speech datasets hinders generalisation.
Only using subsets of the data for training may therefore aid generalisation.

\subsection{Hate Check Evaluation}
\label{app:hate-check}
\begin{table*}[h]
    \resizebox{\textwidth}{!}{
    \renewcommand{\arraystretch}{1.27}
    \begin{tabular}{c|ccc|ccc|ccc|ccc|ccc}
        \multirow{3}{*}{\textbf{Functionality}} & \multicolumn{15}{c}{\textbf{Accuracy} (\%)}\\
        & \multicolumn{3}{c|}{\textbf{Logistic Regression}} & \multicolumn{3}{c|}{\textbf{Bi-LSTM}} & \multicolumn{3}{c|}{\textbf{FNet}} & \multicolumn{3}{c|}{\textbf{DistilBERT}} & \multicolumn{3}{c}{\textbf{RoBERTa}}\\
        & \textbf{Central} & \textbf{F\textsubscript{prox}} & \textbf{F\textsubscript{Opt}} & \textbf{Central} & \textbf{F\textsubscript{prox}} & \textbf{F\textsubscript{Opt}} & \textbf{Central} & \textbf{F\textsubscript{prox}} & \textbf{F\textsubscript{Opt}} & \textbf{Central} & \textbf{F\textsubscript{prox}} & \textbf{F\textsubscript{Opt}} & \textbf{Central} & \textbf{F\textsubscript{prox}} & \textbf{F\textsubscript{Opt}}\\\hline
        
         \textbf{F1}: Expression of strong negative emotions (explicit) & $96.4$ & $\mathbf{100.0}$ & $97.1$ & $80.7$ & $\mathbf{100.0}$ & $95.7$ & $75.0$ & $\mathbf{98.6}$ & $97.9$ & $90.0$ & $\mathbf{99.3}$ & $87.9$ & $89.3$ & $87.9$ & $\mathbf{92.9}$ \\
         \textbf{F2}: Description using very negative attributes (explicit) & $95.0$ & $\mathbf{100.0}$ & $97.9$ & $65.7$ & $99.3$ & $99.3$ & $87.1$ & $\mathbf{100.0}$ & $\mathbf{100.0}$ & $96.4$ & $\mathbf{100.0}$ & $97.7$ & $92.9$ & $93.6$ & $\mathbf{95.0}$ \\
         \textbf{F3}: Dehumanisation (explicit) & $97.9$ & $\mathbf{100.0}$ & $94.3$ & $77.9$ & $\mathbf{100.0}$ & $\mathbf{100.0}$ & $85.0$ & $\mathbf{100.0}$ & $\mathbf{100.0}$ & $97.1$ & $\mathbf{100.0}$ & $93.6$ & $90.7$ & $\mathbf{94.3}$ & $\mathbf{94.3}$ \\
         \textbf{F4}: Implicit derogation & $87.1$ & $\mathbf{95.7}$ & $75.7$ & $70.7$ & $\mathbf{94.3}$ & $92.9$ & $62.1$ & $\mathbf{99.3}$ & $96.4$ & $72.9$ & $\mathbf{82.9}$ & $\mathbf{82.9}$ &  $77.1$ & $78.6$ & $\mathbf{80.7}$ \\\hline
         
         \textbf{F5}: Direct threat & $90.2$ & $\mathbf{99.3}$ & $94.0$ & $80.0$ & $\mathbf{96.2}$ & $88.0$ & $82.0$ & $\mathbf{100.0}$ & $98.5$ & $88.0$ & $\mathbf{98.5}$ & $91.7$ & $91.0$ & $\mathbf{95.5}$ & $91.7$ \\
         \textbf{F6}: Threat as normative statement & $94.3$ & $\mathbf{99.3}$ & $96.4$ & $80.7$ & $\mathbf{99.3}$ & $98.6$ & $70.0$ & $\mathbf{100.0}$ & $\mathbf{100.0}$ & $90.0$ & $\mathbf{100.0}$ & $94.3$ & $\mathbf{96.4}$ & $90.7$ & $91.4$ \\\hline
        
        \textbf{F7}: Hate expressed using slur & $87.5$ & $\mathbf{99.3}$ & $98.6$ & $75.7$ & $88.2$ & $\mathbf{96.5}$ & $86.1$ & $\mathbf{98.6}$ & $96.5$ & $91.0$ & $\mathbf{94.4}$ & $90.0$ & $\mathbf{88.2}$ & $84.7$ & $86.1$ \\
        \textbf{F8}: Non-hateful homonyms of slurs & $6.7$ & $16.7$ & $\mathbf{43.3}$ & $10.0$ & $\mathbf{43.3}$ & $40.0$ & $16.7$ & $23.3$ & $\mathbf{26.7}$ & $23.3$ & $\mathbf{50.0}$ & $33.3$ & $26.7$ & $\mathbf{40.0}$ & $36.7$ \\
        \textbf{F9}: Reclaimed slurs & $4.9$ & $4.9$ & $\mathbf{40.7}$ & $2.5$ & $\mathbf{42.0}$ & $27.2$ & $9.9$ & $11.1$ & $\mathbf{13.6}$ & $6.2$ & $7.4$ & $\mathbf{12.4}$ & $4.9$ & $16.1$ & $\mathbf{17.3}$ \\\hline
         
        \textbf{F10}: Hate expressed using profanity & $\mathbf{100.0}$ & $\mathbf{100.0}$ & $\mathbf{100.0}$ & $93.6$ & $\mathbf{96.4}$ & $\mathbf{96.4}$ & $94.3$ & $\mathbf{100.0}$ & $\mathbf{100.0}$ & $97.9$ & $\mathbf{100.0}$ & $\mathbf{100.0}$ & $\mathbf{100.0}$ & $\mathbf{100.0}$ & $\mathbf{100.0}$ \\
        \textbf{F11}: Non-hateful use of profanity & $2.0$ & $13.0$ & $\mathbf{38.0}$ & $19.0$ & $\mathbf{47.0}$ & $40.0$ & $6.0$ & $11.0$ & $\mathbf{19.0}$ & $15.0$ & $13.0$ & $\mathbf{16.0}$ & $3.0$ & $11.0$ & $\mathbf{17.0}$ \\\hline
        
        \textbf{F12}: Hate expressed through reference in subsequent clauses & $\mathbf{100.0}$ & $\mathbf{100.0}$ & $90.0$ & $91.4$ & $\mathbf{98.6}$ & $\mathbf{98.6}$ & $86.4$ & $\mathbf{100.0}$ & $\mathbf{100.0}$ & $95.0$ & $\mathbf{98.6}$ & $96.4$ & $90.0$ & $\mathbf{92.9}$ & $92.1$ \\
        \textbf{F13}: Hate expressed through reference in subsequent sentences & $\mathbf{100.0}$ & $\mathbf{100.0}$ & $96.2$ & $85.0$ & $\mathbf{96.2}$ & $94.7$ & $84.2$ & $\mathbf{100.0}$ & $\mathbf{100.0}$ & $95.5$ & $\mathbf{99.3}$ & $97.0$ & $\mathbf{96.2}$ & $94.0$ & $95.5$ \\\hline
        
         \textbf{F14}: Hate expressed using negated positive statement & $92.9$ & $\mathbf{99.3}$ & $84.3$ & $57.9$ & $89.3$ & $\mathbf{93.6}$ & $52.1$ & $\mathbf{100.0}$ & $\mathbf{100.0}$ & $77.9$ & $\mathbf{93.6}$ & $76.4$ & $61.4$ & $81.4$ & $\mathbf{90.7}$ \\ 
        \textbf{F15}: Non-hate expressed using negated hateful statement & $6.0$ & $30.0$ & $\mathbf{58.7}$ & $25.6$ & $\mathbf{53.4}$ & $41.4$ & $27.1$ & $23.3$ & $\mathbf{33.8}$ & $17.3$ & $27.1$ & $\mathbf{43.6}$ & $31.6$ & $51.1$ & $\mathbf{56.4}$ \\\hline
         
        \textbf{F16}: Hate phrased as a question & $95.7$ & $\mathbf{100.0}$ & $95.0$ & $81.4$ & $93.6$ & $\mathbf{96.4}$ & $61.4$ & $\mathbf{95.0}$ & $\mathbf{95.0}$ & $92.1$ & $\mathbf{96.4}$ & $91.4$ & $82.1$ & $\mathbf{95.0}$ & $87.9$ \\
        \textbf{F17}: Hate phrased as an opinion & $99.0$ & $\mathbf{100.0}$ & $92.5$ & $89.5$ & $\mathbf{99.0}$ & $97.7$ & $81.2$ & $\mathbf{100.0}$ & $94.0$ & $91.0$ & $\mathbf{98.5}$ & $93.2$ & $86.5$ & $\mathbf{93.2}$ & $87.2$ \\\hline
         
        \textbf{F18}: Neutral statements using protected group identifiers & $20.6$ & $56.3$ & $\mathbf{77.0}$ & $42.1$ & $\mathbf{75.4}$ & $62.7$ & $50.0$ & $55.6$ & $\mathbf{69.0}$ & $69.0$ & $68.3$ & $\mathbf{75.4}$ & $69.0$ & $\mathbf{92.9}$ & $87.3$ \\
        \textbf{F19}: Positive statements using protected group identifiers & $18.0$ & $42.9$ & $\mathbf{80.0}$ & $46.0$ & $\mathbf{64.6}$ & $41.3$ & $45.5$ & $37.0$ & $\mathbf{59.3}$ & $38.6$ & $49.2$ & $\mathbf{73.1}$ & $48.1$ & $78.8$ & $\mathbf{92.1}$ \\\hline
         
        \textbf{F20}: Denouncements of hate that quote it & $1.7$ & $19.0$ & $\mathbf{55.5}$ & $14.5$ & $\mathbf{44.5}$ & $28.3$ & $31.2$ & $\mathbf{45.1}$ & $33.5$ & $16.2$ & $\mathbf{48.6}$ & $38.7$ & $15.6$ & $36.4$ & $\mathbf{37.0}$ \\
        \textbf{F21}: Denouncements of hate that make direct reference to it & $4.2$ & $15.6$ & $\mathbf{47.5}$ & $21.3$ & $\mathbf{46.8}$ & $36.9$ & $27.7$ & $22.7$ & $\mathbf{34.0}$ & $12.1$ & $16.3$ & $\mathbf{30.5}$ & $19.1$ & $39.0$ & $\mathbf{43.3}$ \\\hline
         
        \textbf{F22}: Abuse targeted at objects & $10.8$ & $45.1$ & $\mathbf{70.8}$ & $46.1$ & $\mathbf{70.8}$ & $52.3$ & $53.8$ & $47.7$ & $\mathbf{58.5}$ & $55.4$ & $\mathbf{66.2}$ & $\mathbf{66.2}$ & $60.0$ & $73.8$ & $\mathbf{75.4}$ \\
        \textbf{F23}: Abuse targeted at individuals (not as member of a prot. group) & $4.6$ & $29.2$ & $\mathbf{61.5}$ & $29.2$ & $\mathbf{69.2}$ & $52.3$ & $20.0$ & $24.6$ & $\mathbf{38.5}$ & $20.0$ & $29.2$ & $\mathbf{38.5}$ & $23.1$ & $15.4$ & $\mathbf{69.2}$ \\ 
        \textbf{F24}: Abuse targeted at non-protected groups (e.g. professions) & $14.5$ & $24.2$ & $\mathbf{62.9}$ & $27.4$ & $\mathbf{74.2}$ & $53.2$ & $35.5$ & $40.3$ & $\mathbf{46.8}$ & $41.9$ & $43.5$ & $\mathbf{59.7}$ & $29.0$ & $45.2$ & $\mathbf{66.1}$ \\\hline
         
        \textbf{F25}: Swaps of adjacent characters & $97.7$ & $\mathbf{99.2}$ & $90.2$ & $82.7$ & $\mathbf{95.5}$ & $\mathbf{95.5}$ & $72.2$ & $\mathbf{98.5}$ & $\mathbf{98.5}$ & $78.2$ & $\mathbf{97.0}$ & $78.9$ & $71.4$ & $\mathbf{89.5}$ & $80.5$ \\
        \textbf{F26}: Missing characters & $83.8$ & $\mathbf{100.0}$ & $81.6$ & $72.8$ & $97.1$ & $\mathbf{98.8}$ & $74.6$ & $\mathbf{97.7}$ & $92.4$ & $84.4$ & $94.8$ & $\mathbf{98.8}$ & $89.0$ & $90.2$ & $\mathbf{92.5}$ \\
        \textbf{F27}: Missing word boundaries & $97.9$ & $\mathbf{100.0}$ & $99.2$ & $82.3$ & $\mathbf{100.0}$ & $\mathbf{100.0}$ & $79.4$ & $\mathbf{100.0}$ & $93.6$ & $79.4$ & $\mathbf{90.8}$ & $86.5$ & $\mathbf{93.6}$ & $90.8$ & $90.8$ \\
        \textbf{F28}: Added spaces between chars & $83.8$ & $\mathbf{100.0}$ & $86.1$ & $72.8$ & $97.1$ & $\mathbf{98.8}$ & $74.6$ & $\mathbf{97.7}$ & $92.5$ & $84.4$ & $94.8$ & $\mathbf{98.8}$ & $89.0$ & $90.1$ & $\mathbf{92.5}$ \\
        \textbf{F29}: Leet speak spellings & $99.4$ & $\mathbf{100.0}$ & $98.8$ & $80.9$ & $\mathbf{99.4}$ & $98.8$ & $65.9$ & $\mathbf{98.2}$ & $93.6$ & $75.1$ & $\mathbf{84.4}$ & $75.1$ & $75.1$ & $83.8$ & $\mathbf{86.7}$ \\\hline
    \end{tabular}
    }
    \caption{Results on the HateCheck test suite.}
    \label{tab:hatecheck-table}
    \vspace{-0.2cm}
\end{table*}

This section extends the experiments to qualitatively evaluate the effectiveness of federated and centralised models under different axis of hate speech using \textsc{HateCheck} \cite{rottger-etal-2021-hatecheck}. 
\textsc{HateCheck} is a suite of functional tests for hate speech detection models.
\textsc{HateCheck} provides an in-depth examination of model performances across different potential challenges for machine learning models trained for hate speech detection.

The \textsc{HateCheck} \cite{rottger-etal-2021-hatecheck} dataset consists of $29$ tests, $18$ of which test for distinct expressions of hate while the remaining $11$ test for non-hateful expressions.
The dataset contains $3.728$ labelled samples, $69\%$ of which are `Hate` and while the remaining $31\%$ are labelled as `Not-Hate`. 
We evaluate all the models that have been trained for this manuscript, including the model examined in \cref{sec:appendix}.
We evaluate our trained models \textsc{HateCheck}'s binary form by mapping all classes positive classes to ``hate'' and the negative class to ``not-hate''.\footnote{The \texttt{Comb} dataset uses `none' as its negative class, the Binary Dataset \cite{vidgen-etal-2021-learning} has `Not-hate' as non-hateful label, and Multi-class Dataset~\citet{vidgen-etal-2021-learning} has `None' as non-hateful label}

Conducting the HateCheck functional tests for the models trained on the \texttt{Comb} dataset, we see (please refer to \cref{tab:hatecheck-table}) that the federated learning models perform on par or better than the centralised models on a macro scale. 
The federated Bi-LSTM and FNet models yield strong improvement of 3 - 5\%. 
On the other hand, there is a slight performance dip (0.5 - 1\%) for the federated DistilBERT and RoBERTa models. 
Moreover, through a fine-grained analysis of model performance, we observe that all the models (centralised and federated) perform acceptable performances for different types of derogatory, pronoun reference, phrasing, spelling variations, and threatening language. 
However, all models perform poorly for the tests for counter speech, indicating that while the models learn to recognise some forms of hate, they cannot accurately recognise responses to it.
Furthermore, we see that RoBERTa performs slightly better than all the other model variants on non-hate group identity and abuse against non-protected targets. 
RoBERTa and DistilBERT achieve the best performances for slurs. 
Overall, we find that RoBERTa and DistilBERT consistently perform well across many of the functional tests which might be due to having been pre-trained on large amount of language data. 
However, the pre-training also induces certain biases which limit the models’ performance on profanity. 
The Bi-LSTMs outperform all the models on non-hateful profanity but simultaneously under-perform on hateful profanity.

\section{Conclusion}
Private and sensitive data can risk being exposed when developing and deploying models for hate speech detection.
We therefore examine the use of Federated Learning, a privacy preserving machine learning paradigm to the task of hate speech detection to emphasise privacy in hate speech detection.
We find that using Federated Learning improves on the performance levels achieved using centralised models, thus affording both privacy and performance.
In future work, we intend to examine interpretability and explainability for federated learning to gain a better understanding of the causes of such performance increases.

\section*{Limitations}
While Federated Learning introduces increased privacy in the process of hate speech detection, a real time system may be vulnerable to attacks that can lead to privacy leakages. 
For instance, the weights being transferred from the clients to the server may reveal information about the local dataset to an adversary~\citep{bhowmick2018protection,melis2019exploiting}. 
However unintended these leakages may be, they still pose a significant threat and might limit the privacy claim.

The Federated Learning models trained in our work rely on $8$ of the $9$ datasets used by \citet{fortuna2021well}, as we could not gain access to the final dataset. 
We do not test the biases introduced in Federated Learning models upon combining and normalising these datasets under the schema proposed by \citet{fortuna2021well,fortuna-etal-2020-toxic}. 
Additionally, the dataset division for the simulation is done under the assumption of I.I.D. conditions which might not always be true for real-world scenarios.

\section*{Ethical Considerations}

Although our methods for hate speech detection provide increased privacy to downstream users of content moderation technologies, i.e. users of online platforms, there are significant risks to it.
First, our proposed technology has dual use implications, as it can also be applied maliciously, for instance to limit the speech of specific groups.
Second, while this work uses publicly available datasets, there is an inherent tension between the public availability of data and privacy risks.
Finally, although all model updates occur on local client devices, federated learning is not a silver bullet which addresses issues of systemic violence of content moderation \citet{thylstrup_detecting_2020}, or issues of privacy.
Rather, federated learning can provide an avenue for engaging in meaningful conversations with people and their experiences and needs for content moderation and privacy.

\bibliography{anthology,custom,references}
\bibliographystyle{acl_natbib}

\appendix
\section{Learning From the Worst}
\label{sec:appendix}
Extending the experiments conducted in~\cref{sec:experiments}, we aim to analyse if our claims are corroborated when we expose the complete setup of federated as well as centralised models to other datasets.
We perform this analysis on the ``Learning from the Worst'' dataset \cite{vidgen-etal-2021-learning}.

\subsection{Dataset}
\label{sec:appendix-data}


\paragraph{Binary Dataset} We use the Dynamically Generated Hate Dataset v0.2.2 provided by~\citet{vidgen-etal-2021-learning} which contains $41,255$ entries. 
We use the training, testing, and validation sets provided by~\citet{vidgen-etal-2021-learning}. 
This dataset consists of two categories: hate and not-hate. The category distribution is shown in~\cref{category-table-zeerak}.

\paragraph{Multi-class Dataset} We use the same Dynamically Generated Hate Dataset v0.2.2 provided by~\citet{vidgen-etal-2021-learning}. However, we make use of the multi-class labels provided in the original dataset. 
It consists of seven categories: none (i.e. not-hate), derogation, not-given, animosity, dehumanisation, threatening, and support (see \cref{category-table-zeerak} for class distribution).

\begin{table}[H]
\centering
\resizebox{\columnwidth}{!}{
\begin{tabular}{c|cc}

{\textbf{Binary categories}}  &
{\textbf{Multi-class categories}} &
{\textbf{Count}} \\\hline

Not-hate & None & $18,993$ \\\hline
\multirow{6}{*}{Hate} & Derogation & $9,907$ \\ 
& Not Given & $7,197$ \\ 
& Animosity & $3,439$ \\ 
& Dehumanisation & $906$ \\ 
& Threatening & $606$ \\ 
& Support & $207$ \\

\end{tabular}
}
\caption{Label distribution of \citet{vidgen-etal-2021-learning} v.0.2.2.}
\label{category-table-zeerak}
\end{table}

\begin{table}[H]

\resizebox{\columnwidth}{!}{
\begin{tabular}{l|ccc|ccc}

\textbf{Model}
& \multicolumn{3}{c|} {\textbf{Binary Dataset}}
& \multicolumn{3}{c} {\textbf{Multi-class Dataset}} \\

& Precision & Recall & F1 
& Precision & Recall & F1  \\\hline 

LogReg
& $63.38$ & $52.58$ & $54.98$
& $56.68$ & $35.46$ & $40.92$  \\

Bi-LSTM 
& $63.56$ & $52.90$ & $55.38$ 
& $58.44$ & $41.70$ & $45.91$\\

FNet 
& $27.75$ & $48.05$ & $33.74$
& $53.41$ & $23.46$ & $27.60$\\

DistilBERT 
& $71.56$ & $71.72$ & $68.63$ 
& $78.25$ & $52.96$ & $59.27$\\

RoBERTa 
& $\textbf{76.18}$ & $\textbf{77.50}$ & $\textbf{74.37}$ 
& $\textbf{80.26}$ & $\textbf{62.69}$ & $\textbf{67.91}$\\

\end{tabular}
}
\caption{Centralised model performances for binary and multi-class datasets}
\label{tab:server-results-zeerak}

\end{table}

\begin{table*}[h]
    \resizebox{\textwidth}{!}{
    \centering
    \begin{tabular}{ccc|ccc|ccc|ccc|ccc}
        ~ & ~ & ~ & \multicolumn{3}{c|}{Binary (FedProx)} & \multicolumn{3}{c|}{Binary (FedOpt)} & \multicolumn{3}{c|}{Multiclass (FedProx)} & \multicolumn{3}{c}{Multiclass (FedOpt)}\\
        ~ & ~ & ~ & \textbf{Precision} & \textbf{Recall} & \textbf{F1} & \textbf{Precision} & \textbf{Recall} & \textbf{F1} & \textbf{Precision} & \textbf{Recall} & \textbf{F1} & \textbf{Precision} & \textbf{Recall} & \textbf{F1}\\\hline

        ~ & ~ & \textbf{e = 1} & 59.43 & 59.52 & 59.34 & 57.38 & 57.44 & 57.36 & 45.55 & 39.62 & 41.34 & 45.06 & 37.56 & 40.15\\ 
        ~ & \textbf{c = 10\%} & \textbf{e = 5} & 60.13 & 59.99 & 60.01 & 55.93 & 55.79 & 55.76 & 47.03 & 44.71 & 45.58 & 44.93 & 43.91 & 44.25\\ \ 
        ~ & ~ & \textbf{e = 20} & 60.08 & 59.94 & 59.97 & 56.08 & 55.98 & 55.97 & 47.68 & 48.01 & 47.77 & 46.17 & 46.39 & 46.11\\ 
        Logistic  & ~ & \textbf{e = 1} & 59.78 & 59.69 & 59.71 & 55.88 & 55.74 & 55.71 & 47.28 & 34.94 & 38.20 & 43.26 & 36.66 & 39.07\\ 
        Regression & \textbf{c = 30\%} & \textbf{e = 5} & 60.79 & 60.80 & 60.80 & 55.11 & 55.02 & 55.01 & 48.12 & 47.58 & 47.74 & 44.77 & 42.05 & 43.17\\ 
        ~ & ~ & \textbf{e = 20} & 60.76 & 60.41 & 60.41 & 54.93 & 54.80 & 54.76 & 47.27 & 47.94 & 47.50 & 45.26 & 46.44 & 45.51\\ 
        ~ & ~ & \textbf{e = 1} & 60.60 & 60.55 & 60.57 & 54.46 & 54.36 & 54.33 & 46.92 & 34.00 & 37.08 & 42.78 & 36.52 & 38.44\\ 
        ~ & \textbf{c = 50\%} & \textbf{e = 5} & 60.76 & 60.73 & 60.74 & 54.58 & 54.52 & 54.51 & 47.54 & 46.68 & 46.96 & 43.60 & 42.26 & 42.85\\ 
        ~ & ~ & \textbf{e = 20} & 60.79 & 60.55 & 60.57 & 54.85 & 54.72 & 54.66 & 47.55 & 48.91 & 48.10 & 44.28 & 44.93 & 44.41\\\hline

        ~ & ~ & \textbf{e = 1} & 61.15 & 61.26 & 61.00 & 61.05 & 61.17 & 60.96 & 38.54 & 38.02 & 35.96 & 40.16 & 38.75 & 38.23\\ 
        ~ & \textbf{c = 10\%} & \textbf{e = 5} & 57.63 & 57.55 & 57.57 & 57.93 & 57.88 & 57.89 & 46.09 & 45.68 & 45.84 & 45.88 & 43.57 & 44.34\\ 
        ~ & ~ & \textbf{e = 20} & 58.15 & 58.14 & 58.15 & 58.70 & 58.65 & 58.66 & 45.24 & 47.79 & 45.92 & 45.49 & 49.88 & 45.94\\ 
        ~ & ~ & \textbf{e = 1} & 61.72 & 61.74 & 61.13 & 60.26 & 60.37 & 60.13 & 40.23 & 36.64 & 34.89 & 43.64 & 31.18 & 32.74\\ 
        Bi-LSTM & \textbf{c = 30\%} & \textbf{e = 5} & 58.48 & 58.48 & 58.48 & 59.36 & 59.42 & 59.36 & 45.77 & 44.33 & 44.97 & 46.30 & 43.19 & 44.00\\ 
        ~ & ~ & \textbf{e = 20} & 57.07 & 57.06 & 57.06 & 59.37 & 59.35 & 59.36 & 45.06 & 47.07 & 45.58 & 46.68 & 50.00 & 47.19\\ 
        ~ & ~ & \textbf{e = 1} & 60.73 & 60.84 & 60.66 & 60.50 & 60.58 & 60.20 & 45.05 & 32.61 & 34.86 & 43.98 & 33.39 & 35.40\\ 
        ~ & \textbf{c = 50\%} & \textbf{e = 5} & 57.90 & 57.93 & 57.91 & 59.51 & 59.55 & 59.52 & 45.59 & 44.88 & 45.22  & 46.85 & 45.15 & 45.77\\ 
        ~ & ~ & \textbf{e = 20} & 57.30 & 57.28 & 57.29 & 59.26 & 59.25 & 59.25 & 45.50 & 48.40 & 46.00  & 46.73 & 50.02 & 47.37\\\hline

        ~ & ~ & \textbf{e = 1} & 72.84 & 71.99 & 72.16 & 73.13 & 70.57 & 70.65 & 38.33 & 32.35 & 25.99 & 39.64 & 33.78 & 23.90\\ 
        ~ & \textbf{c = 10\%} & \textbf{e = 5} & 69.48 & 69.58 & 69.51 & 69.93 & 69.10 & 69.22 & 54.00 & 54.95 & 54.11 & 40.93 & 26.84 & 17.58\\ 
        ~ & ~ & \textbf{e = 20}  & 69.84 & 68.99 & 69.11 & 70.03 & 69.97 & 69.99 & 50.80 & 52.13 & 51.37 & 49.16 & 50.48 & 49.39\\ 
        ~ & ~ & \textbf{e = 1} & 72.48 & 71.76 & 71.91 & 72.31 & 71.53 & 71.69 & 53.24 & 39.65 & 40.27 & 51.74 & 35.40 & 37.36\\ 
        FNet & \textbf{c = 30\%} & \textbf{e = 5}  & 70.59 & 70.29 & 70.38 & 71.47 & 70.72 & 70.87 & 53.02 & 52.47 & 51.62 & 43.53 & 28.59 & 22.33\\ 
        ~ & ~ & \textbf{e = 20}  & 69.06 & 68.93 & 68.98 & 69.96 & 69.93 & 69.94 & 51.03 & 52.98 & 51.43 & 49.66 & 52.62 & 49.74\\ 
        ~ & ~ & \textbf{e = 1}  & 72.64 & 72.47 & 72.54 & 72.61 & 72.47 & 72.53 & 58.49 & 40.30 & 43.27 & 49.48 & 32.74 & 33.23\\ 
        ~ & \textbf{c = 50\%} & \textbf{e = 5}  & 69.67 & 69.28 & 69.39 & 70.19 & 69.92 & 70.00 & 54.31 & 53.25 & 53.54 & 50.44 & 50.07 & 49.82\\ 
        ~ & ~ & \textbf{e = 20}  & 67.71 & 67.62 & 67.66 & 68.21 & 68.14 & 68.17 & 51.62 & 51.75 & 51.55 & 48.10 & 50.00 & 48.27\\\hline

        ~ & ~ & \textbf{e = 1} & 74.62 & 74.76 & 74.67 & 74.20 & 74.43 & 74.21 & 58.99 & 42.73 & 45.33 & 63.91 & 49.15 & 52.88\\ 
        ~ & \textbf{c = 10\%} & \textbf{e = 5} & 73.74 & 73.05 & 73.21 & 72.35 & 72.17 & 72.24 & 60.72 & 58.22 & 59.27 & 60.41 & 57.57 & 58.71\\ 
        ~ & ~ & \textbf{e = 20} & 72.92 & 72.39 & 72.53 & 70.77 & 70.43 & 70.53 & 59.61 & 59.14 & 59.24 & 58.78 & 59.38 & 58.95\\ 
        ~ & ~ & \textbf{e = 1} & 74.90 & 74.77 & 74.82 & 73.73 & 73.84 & 73.78 & 58.95 & 43.48 & 45.98 & 54.13 & 40.11 & 39.49\\ 
        DistilBERT & \textbf{c = 30\%} & \textbf{e = 5} & 73.34 & 73.14 & 73.22 & 70.98 & 70.89 & 70.93 & 60.66 & 58.46 & 59.39 & 60.73 & 58.97 & 59.72\\ 
        ~ & ~ & \textbf{e = 20} & 73.02 & 72.82 & 72.90 & 70.37 & 70.17 & 70.24 & 58.79 & 58.85 & 58.71 & 58.22 & 58.80 & 58.28\\ 
        ~ & ~ & \textbf{e = 1} & 74.94 & 74.88 & 74.91 & 73.60 & 73.72 & 73.64 & 59.74 & 43.91 & 46.44 & 55.35 & 40.25 & 41.26\\ 
        ~ & \textbf{c = 50\%} & \textbf{e = 5} & 73.35 & 73.02 & 73.13 & 70.51 & 70.23 & 70.32 & 60.52 & 58.39 & 59.26 & 60.35 & 59.26 & 59.71\\ 
        ~ & ~ & \textbf{e = 20} & 73.07 & 72.67 & 72.79 & 70.08 & 69.84 & 69.92 & 58.85 & 58.39 & 58.54 & 58.41 & 58.60 & 58.36\\\hline
        
        ~ & ~ & \textbf{e = 1} & 80.72 & 80.90 & 80.78 & 80.74 & 81.01 & 80.46 & 62.50 & 48.67 & 51.76 & 66.26 & 54.58 & 58.04\\ 
        ~ & \textbf{c = 10\%} & \textbf{e = 5} & 80.97 & 80.94 & 80.95 & 80.20 & 80.38 & 80.27 & 65.33 & 64.86 & 64.92 & 63.97 & 62.20 & 62.93 \\ 
        ~ & ~ & \textbf{e = 20} & 81.71 & 81.82 & 81.76 & 80.34 & 80.40 & 80.37 & 63.74 & 64.28 & 63.85 & 63.68 & 63.80 & 63.65\\ 
        ~ & ~ & \textbf{e = 1} & 81.61 & 81.76 & 81.67 & 80.79 & 80.98 & 80.86 & 64.77 & 47.26 & 50.49 & 65.87 & 58.10 & 60.41\\ 
        RoBERTa & \textbf{c = 30\%} & \textbf{e = 5} & 81.27 & 81.33 & 81.30 & 79.92 & 80.10 & 79.98 & 65.27 & 64.31 & 64.64 & 63.58 & 63.19 & 63.22\\ 
        ~ & ~ & \textbf{e = 20} & 81.22 & 81.37 & 81.28 & 79.17 & 79.32 & 79.23 & 64.59 & 64.82 & 64.54 & 63.14 & 63.16 & 62.96\\ 
        ~ & ~ & \textbf{e = 1} & 80.82 & 80.81 & 80.82 & 80.84 & 81.09 & 80.91 & 64.81 & 45.97 & 49.10 & 66.27 & 58.58 & 61.14\\ 
        ~ & \textbf{c = 50\%} & \textbf{e = 5} & 81.76 & 81.86 & 81.80 & 79.66 & 79.75 & 79.70 & 65.19 & 64.50 & 64.68 & 64.13 & 63.77 & 63.82\\ 
        ~ & ~ & \textbf{e = 20} & 81.32 & 81.41 & 81.36 & 79.85 & 79.94 & 79.89 & 64.83 & 64.62 & 64.56  & 63.54 & 63.19 & 63.29 \\ 
    \end{tabular}
    }
    \caption{Results of binary and multi-class classification experiments run on the datasets released by~\citet{vidgen-etal-2021-learning} using the Federated Learning setup (FedProx and FedOpt). c is the percentage of clients whose updates are considered. e is the number of local epochs on edge device.}
\label{tab:fed-multi-results}
\end{table*}

\begin{table}
\end{table}

\begin{table}[ht]
\centering
\resizebox{\columnwidth}{!}{
\begin{tabular}{l|cc|cc}

\textbf{Model} 
& \multicolumn{2}{c|} {\textbf{Binary Dataset}}
& \multicolumn{2}{c} {\textbf{Multi-class Dataset}} \\

& Server trained & Federated
& Server trained & Federated  \\\hline

LogReg
& $59.61$ & $60.74$
& $39.79$ & $48.10$\\

Bi-LSTM 
& $61.98$ & $61.13$
& $42.18$ & $47.37$\\

FNet 
& $35.51$ & $72.54$
& $45.52$ & $54.11$\\

DistilBERT 
& $74.95$ & $74.91$ 
& $59.76$ & $59.72$\\

RoBERTa 
& $80.64$ & $81.80$ 
& $65.73$ & $64.92$\\

\end{tabular}
}
\caption{Results comparing the F1 scores for server-based approaches and federated approaches for binary and multi-class datasets proposed by \newcite{vidgen-etal-2021-learning}}
\label{tab:result-comparison-zeerak}
\end{table}

\subsection{Analysis}
\label{sec:appendix-analysis}
We follow the training procedures outlined in \cref{sec:experiments} on the binary and multi-class versions of the~\citet{vidgen-etal-2021-learning} dataset and consider the results on the multi-class dataset (see \Cref{tab:server-results-zeerak,tab:fed-multi-results}).
We observe similar performance trends for the Logistic Regression and Bi-LSTM models in \cref{tab:result-comparison-zeerak} to those for \texttt{Comb}. 
This pattern extends to the Transformer-based models with the exception of the RoBERTa model.
The federated RoBERTa obtains a slightly lower F1 score than the centralised version in ($64.92$ and $65.73$, respectively).

The pattern of performances for the binary dataset varies from our main dataset. 
Here, we observe in \Cref{tab:fed-multi-results} that the FNet model adapts well to the federated setting, with both FedProx and FedOpt algorithms significantly improving on their centralised counter-part ($65.14$ and $35.51$, respectively) (see ~\Cref{tab:result-comparison-zeerak}).
Moreover, we find that the models optimised using FedProx algorithm outperform those using the FedOpt algorithm for all federated learning settings with the exception of the DistilBERT variant with $c = 50\%$ and $e = 5$. 
For the binary dataset, we observe that all federated models except for RoBERTa perform better across client fractions when trained for lower epochs. For the multi-class dataset, however, all federated models have improved performance across client fractions, when the models are trained for a higher number of epochs.
We observe from our results that there is a slight performance decrease for the federated versions of the Bi-LSTM, RoBERTa, and DistilBERT, when compared to the centralised models.
A small decrease in performance however is expected for federated learning, due to its emphasis on privacy protections.
In spite of small differences, the experiments on both the binary and multi-class versions of \newcite{vidgen-etal-2021-learning} closely resemble the results obtained on \texttt{Comb}, suggesting that federated learning is applicable across datasets for hate speech and class distributions.

We see a similar trend as \texttt{Comb} while performing the \textsc{HateCheck} functional tests on the binary and multi-class dataset. 
The Logistic Regression adapts poorly across different types of counter speech, slurs, non-hate group identity, negation, and abuse against non-protected targets. 
Moreover, we also observe that Bi-LSTM  and FNet yield poor performance for different types of negation and non-hate group identity. 
We find that in most cases, models trained on the binary dataset achieve higher performances than the models trained on its multi-class counterpart. 

\section{Model Exploration}
This section highlights the different model settings and hyper-parameter selection strategies used while training the models on \texttt{Comb} and \newcite{vidgen-etal-2021-learning} in \cref{sec:appendix}. 
We also provide a token level analysis conducted by on \texttt{Comb}.

\subsection{Hyper-parameter Search}
\label{app:hyper-params}

We use Weights and Biases~\citep{wandb} as our experiment tracking tool for all experiments. 
We run a Bayesian search for finding the optimal client learning rate, server-side learning rate, and the proximal term. 
In our hyper-parameter search for the value of proximal term, we conduct a categorical search.
Following \newcite{li2018federated}, we set the possible values to $0.001$, $0.01$, $0.1$, and $1$.

\subsection{Model Descriptions}
\label{sec:appendix-fed}
We implement all models using PyTorch~\citep{Paszke_et_al} and Huggingface libraries~\citep{wolf-etal-2020-transformers}. 
We train the Logistic Regression and Bi-LSTM models for $300$ rounds, and transformer-based models for $50$ rounds.
We implement early stopping based on the weighted validation F1 scores, with the patience set to $10$ rounds.
After conducting our hyper-parameter search, we choose our hyper-parameters (see ~\cref{tab:fedprox-configs} and ~\cref{tab:fedopt-configs}).
The measure the performances of all our models using precision, recall and weighted F1 scores.

\subsection{Token Level Analysis}
\label{sec:appendix-token-dist}

\begin{table*}[h]
\centering
\begin{tabular}{ccccc}

{\textbf{Dataset}}  &
{\textbf{Tokenization}} &
{\textbf{Minimum}} &
{\textbf{99\%ile}} &
{\textbf{Maximum}} \\\hline

\multirow{2}{*}{\newcite{talat_hateful_2016}} & Word-level & $1$ & $34$ & $54$ \\ 
& Subword-level & $3$ & $60$ & $101$ \\\hline
\multirow{2}{*}{\newcite{davidson_automated_2017}} & Word-level & $1$ & $37$ & $94$ \\ 
& Subword-level & $2$ & $83$ & $412$ \\\hline
\multirow{2}{*}{\newcite{fersini2018overview}} & Word-level & $2$ & $36$ & $47$ \\ 
& Subword-level & $3$ & $64$ & $93$ \\\hline
\multirow{2}{*}{\newcite{de-gibert-etal-2018-hate}} & Word-level & $1$ & $67$ & $374$ \\ 
& Subword-level & $1$ & $93$ & $592$ \\\hline
\multirow{2}{*}{\newcite{swamy-etal-2019-studying}} & Word-level & $1$ & $175$ & $1481$ \\ 
& Subword-level & $1$ & $233$ & $3209$ \\\hline
\multirow{2}{*}{\newcite{basile-etal-2019-semeval}} & Word-level & $1$ & $59$ & $74$ \\ 
& Subword-level & $3$ & $105$ & $156$ \\\hline
\multirow{2}{*}{\newcite{zampieri-etal-2019-predicting}} & Word-level & $2$ & $69$ & $112$ \\ 
& Subword-level & $4$ & $152$ & $221$ \\\hline
\multirow{2}{*}{Kaggle} & Word-level & $1$ & $727$ & $4950$ \\ 
& Subword-level & $2$ & $872$ & $4952$ \\\hline

\end{tabular}
\caption{Word-level and subword-level (BPE) token sequence length distribution for \texttt{Comb} dataset described in Section \ref{combined-dataset}}
\label{token-seq-length-combined-dataset}
\end{table*}

\begin{figure}[H]
\centering
\includegraphics[width=\columnwidth]{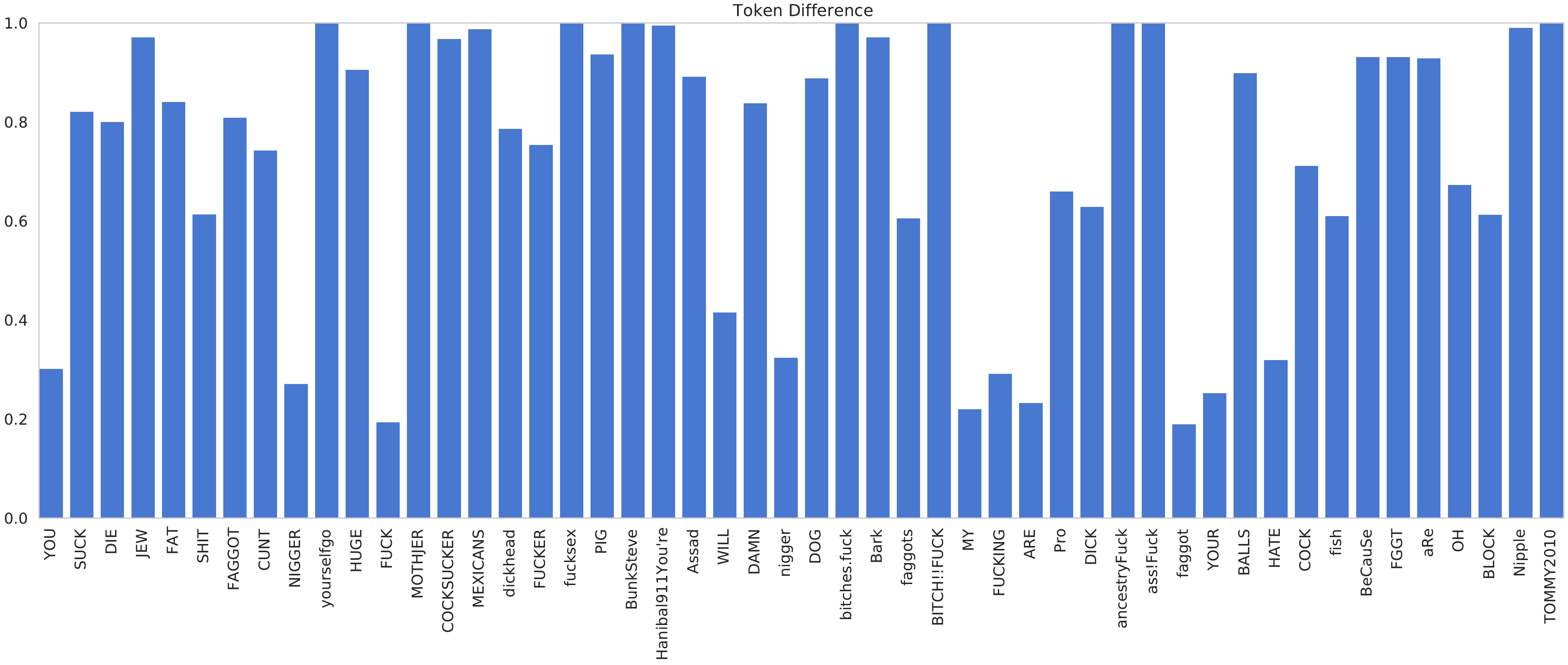}
\caption{fraction of token occurrences (of 50 most frequent tokens) in the discarded data} 
\label{token-fig}
\end{figure}

\Cref{token-seq-length-combined-dataset} shows the preliminary explorations of the token-level distributions for the combined dataset (\Cref{combined-dataset}) with two tokenisation methods: word-level using SpaCy \cite{honnibal_spacy1_2017} 
and subword-level using the BPE algorithm.\footnote{https://github.com/VKCOM/YouTokenToMe}
Based on our analysis, we draw the following conclusions:
1) token length is highly imbalanced for different datasets in \texttt{Comb}, particularly in Kaggle dataset\textsuperscript{\ref{kaggle-data}};
2) 99th percentile token length in Kaggle dataset\textsuperscript{\ref{kaggle-data}} is reflected in the remaining dataset. 
Considering this, we remove longest $1\%$ of documents from the Kaggle dataset\textsuperscript{\ref{kaggle-data}} to achieve faster computation.
Through this exclusion process, the maximal token length of documents is reduced from $4950$ to $727$ tokens, without a substantial loss of information.

\begin{table*}[!ht]

\centering
\begin{tabular}{l|ccc|ccc|ccc}

\textbf{Model} 
& \multicolumn{3}{c|} {\textbf{Combined Datasets}} 
& \multicolumn{3}{c|} {\textbf{Binary Dataset}}
& \multicolumn{3}{c} {\textbf{Multi-class Dataset}} \\

& bs & client\_lr & $\mu$
& bs & client\_lr & $\mu$ 
& bs & client\_lr & $\mu$  \\\hline

LogReg
& $128$ & $0.01$ & $0.01$ 
& $64$ & $0.01$ & $0.01$ 
& $64$ & $0.01$ & $0.01$   \\

Bi-LSTM 
& $128$ & $0.001$ & $0.01$ 
& $64$ & $0.001$ & $0.01$ 
& $64$ & $0.001$ & $0.01$\\

FNet 
& $32$ & $0.0001$ & $0.1$ 
& $32$ & $0.0001$ & $0.001$
& $32$ & $0.0001$ & $0.001$\\

DistilBERT 
& $32$ & $0.00004$ & $0.01$
& $32$ & $0.00002$ & $0.01$ 
& $32$ & $0.00002$ & $0.01$\\

RoBERTa 
& $16$ & $0.00002$ & $0.01$ 
& $24$ & $0.00002$ & $0.01$ 
& $32$ & $0.00002$ & $0.01$\\

\end{tabular}
\caption{Model hyper-parameters for server-based and federated models for the \newcite{vidgen-etal-2021-learning}. `bs' represents batch size, `client\_lr' represents client learning rate, $\mu$ represents proximal term for FedProx algorithm.}
\label{tab:fedprox-configs}

\vspace{5mm}
\end{table*}

\begin{table*}[!ht]

\centering
\begin{tabular}{l|ccc|ccc|ccc}

\textbf{Model} 
& \multicolumn{3}{c|} {\textbf{Combined Datasets}} 
& \multicolumn{3}{c|} {\textbf{Binary Dataset}}
& \multicolumn{3}{c} {\textbf{Multi-class Dataset}} \\

& bs & client\_lr & server\_lr
& bs & client\_lr & server\_lr
& bs & client\_lr & server\_lr  \\\hline

LogReg
& $128$ & $0.01$ & $0.01$
& $64$ & $0.01$ & $0.001$ 
& $64$ & $0.01$ & $0.01$  \\

Bi-LSTM 
& $128$ & $0.001$ & $0.01$ 
& $64$ & $0.001$ & $0.001$ 
& $64$ & $0.001$ & $0.001$\\

FNet 
& $32$ & $0.0001$ & $0.001$ 
& $32$ & $0.0001$ & $0.0001$
& $32$ & $0.0001$ & $0.0001$\\

DistilBERT 
& $32$ & $0.00004$ & $0.001$
& $32$ & $0.00002$ & $0.0001$ 
& $32$ & $0.00002$ & $0.0001$\\

RoBERTa 
& $16$ & $0.00002$ & $0.001$ 
& $24$ & $0.00002$ & $0.0001$ 
& $32$ & $0.00002$ & $0.0001$\\

\end{tabular}
\caption{Model hyper-parameters for server-based and federated models for the \newcite{vidgen-etal-2021-learning}. `bs' represents batch size, `client\_lr' represents client learning rate, `server\_lr' represents server learning rate for FedOpt algorithm.}
\label{tab:fedopt-configs}

\vspace{5mm}
\end{table*}

\end{document}